\def\ARXIV{1} 

\def\APPENDIX{0} 

\documentclass{ecai} 

\usepackage{latexsym}
\usepackage{amssymb}
\usepackage{amsmath}
\usepackage{amsthm}
\usepackage{booktabs}
\usepackage{enumitem}
\usepackage{graphicx}
\usepackage{color}

\usepackage[hidelinks]{hyperref}
\usepackage{multirow} 
\usepackage{multicol} 
\usepackage{orcidlink}
\usepackage{balance}
\usepackage{doi}
\usepackage{caption}
\newcommand{\BibTeX}{B\kern-.05em{\sc i\kern-.025em b}\kern-.08em\TeX}

\PassOptionsToPackage{colorlinks=false}{hyperref}

\ifnum\ARXIV=1
    \usepackage{fancyhdr} 
\fi

\begin{document}


\begin{frontmatter}


\paperid{6025} 


\title{Cross-Modal Temporal Fusion for Financial Market Forecasting}




\author[A]{\fnms{Yunhua}~\snm{Pei}\orcidlink{0000-0003-2906-0827}}
\author[B]{\fnms{John}~\snm{Cartlidge}\orcidlink{0000-0002-3143-6355}\thanks{Corresponding Author. Email: john.cartlidge@bristol.ac.uk.}}
\author[C]{\fnms{Anandadeep}~\snm{Mandal}\orcidlink{0000-0002-1018-8719}}
\author[D]{\fnms{Daniel}~\snm{Gold}}
\author[D]{\fnms{Enrique}~\snm{Marcilio}}
\author[D]{\fnms{Riccardo}~\snm{Mazzon}}

\address[A]{School of Computer Science, University of Bristol, Bristol, UK}
\address[B]{School of Engineering Mathematics and Technology, University of Bristol, Bristol, UK}
\address[C]{Business School, University of Birmingham, Birmingham, UK}
\address[D]{Stratiphy Limited, London, UK}

\begin{abstract}
Accurate forecasting in financial markets requires integrating diverse data sources, from historical prices to macroeconomic indicators and financial news. However, existing models often fail to align these modalities effectively, limiting their practical use. In this paper, we introduce a transformer-based deep learning framework, Cross-Modal Temporal Fusion (CMTF), that fuses structured and unstructured financial data for improved market prediction. The model incorporates a tensor interpretation module for feature selection and an auto-training pipeline for efficient hyperparameter tuning. Experimental results using FTSE 100 stock data demonstrate that CMTF achieves superior performance in price direction classification compared to classical and deep learning baselines. These findings suggest that our framework is an effective and scalable solution for real-world cross-modal financial forecasting tasks.
\end{abstract}
\end{frontmatter}


\section{Introduction}
Forecasting financial markets is a challenging and high-risk task, with implications for investment strategies, risk management, and economic policy. The primary objective is to accurately predict the prices of financial assets in order to generate potential profits. In this context, stock prediction, a crucial aspect of financial markets, has gained increasing attention over the past few years.


The Efficient Market Hypothesis (EMH) \cite{fama1965behavior,fama1970efficient-markets} suggests that market efficiencies place limitations on the ability to consistently generate excess returns. In weak form efficiency, it is assumed that asset prices incorporate all information in past prices, making technical analysis ineffective; in semi-strong form, prices are assumed to reflect all public information, including historical prices, news, earnings reports, and economic data, therefore also rendering fundamental analysis ineffective; and in strong form efficiency prices are assumed to reflect both public and private information, making even insider trading ineffective. However, the EMH is controversial and there is ample empirical evidence of market inefficiencies \cite{barberis2003survey}, suggesting that it is possible to predict prices for excess returns.

The financial industry has been exploring prediction models since the early twentieth century \cite{cowles1933can}, continuously advancing these technologies through substantial financial investments. Traditional quantitative approaches mainly rely on historical time series data to forecast stock movements \cite{poon2003forecasting, taylor2008modelling}. However, with the development of deep learning, more recent efforts have explored approaches to decompose complex market dynamics \cite{zhou2019emd2fnn, pei2025dynamic-icaart} and capture stock interdependencies through attention mechanisms \cite{ijcai2019p514, cui2023temporal}.

Lately, advances in Natural Language Processing (NLP) have enabled the integration of unstructured textual data to enhance prediction models. For example, news \cite{hu2018listening, picasso2019technical, xu2018stock} and social media content \cite{jin2020stock, tolstikhin2021mlp} can be analyzed for sentiment to generate a scoring matrix of positive-negative signals for each stock, which is then incorporated as a new input feature. These event-driven methods focus on extracting valuable patterns from event information for stock prediction.

Although prior work has achieved some success in stock prediction, three open challenges remain: (1) {\em Heterogeneous data integration} -- existing methods \cite[e.g.,][]{poon2003forecasting, zhou2019emd2fnn} tend to crudely aggregate multifrequency inputs (e.g., quarterly reports, daily price series, and real-time news) without aligning their temporal dependencies, which can lead to loss of signal or spurious correlations; (2) {\em Superficial interpretability} -- existing use of attention mechanisms \cite{zhou2020domain, lin2021learning} fails to disentangle specific drivers (e.g., GDP trends vs. news) or provide actionable insights, making it difficult for practitioners to understand or trust predictions; (3) {\em Inflexible training paradigms} -- characterized by rigid architectures with inefficient retraining and hyperparameter optimization \cite{zu2023finformer, choi2023hybrid}, existing methods have limited ability to rapidly adapt to volatile market conditions.



To address these challenges and solve the core problem of aligning and extracting value from diverse financial signals, we propose CMTF, a cross-modal temporal fusion unified framework that (i) integrates multimodal data, (ii) ensures forecasting interpretability, and (iii) automates training schemes for rapid iteration with hyperparameter tuning. The CMTF framework provides valuable insight for practitioners dealing with diverse data types, offering guidance on how to effectively handle and select variable features during the feature engineering process. Furthermore, to improve training efficiency, it introduces hyperparameter search rules to the attention-based model, enabling faster convergence, faster iterations, and optimized performance. The main contributions of this work can be summarized as follows:

\begin{itemize}
    \item We propose a multimodal tensor representation that integrates structured data (historical market data and macro-index) and unstructured data (news sentiment and financial reports); enabling systematic alignment of heterogeneous temporal and event-driven signals for stock market forecasting.
    \item We design a sparse tensor interpretation framework that leverages Lasso regression for feature selection and attention mechanisms to prioritize cross-modal interactions (e.g., linking event-driven trends to modality-specific price movements); ensuring interpretable and actionable predictions.
    \item We conduct extensive experiments on real-world stocks from the FTSE 100 index. The experimental results demonstrate that CMTF outperforms a suite of baselines in forecasting the next trading day close price, with average improvements of 1.52\% in precision, 30.38\% in recall, and 0.17 in F1 score for the classification task.
\end{itemize}

The contributions and findings of this work have informed the development of Stratiphy's emerging applications. Stratiphy is a wealth management platform for retail investors to build active portfolios using industry-leading trading strategies and risk management tools.\footnote{Stratiphy can be found at \url{https://www.stratiphy.io}.} The CMTF framework is being prototyped as an emerging application to develop new investment strategies for business and retail customers. These advances are essential to the continued success of Stratiphy and offer the social benefits of better financial investment and risk management for all. Code and data availability\footnote{Code is available at \url{https://github.com/PEIYUNHUA/CMTF}; for data access, please contact Stratify.}.

\section{Literature Review}
\subsection{Multimodality}
Recent advances in multimodal learning have enabled the fusion of structured and unstructured data for financial forecasting. \cite{lim2021temporal} introduced temporal fusion transformers to jointly model static covariates (e.g., sector metadata) and dynamic time series data, but their fixed temporal alignment struggles with low-frequency earnings reports. To address this, \cite{mandal2024enhancing} introduced a cross-modal transformer to align daily X (formerly Twitter) data with historical price trends, but their fixed temporal windows ignore intermodal frequency mismatches.  A notable advancement is \cite{wood2021trading}, which introduced the momentum transformer, which combines technical indicators with attention mechanisms to capture momentum-driven market regimes. However, like \cite{tavakoli2025multi}, which fused numerical and textual datasets using cross-modal attention, these methods lack mechanisms to synchronize different granularity data with momentum shifts. Furthermore, despite their predictive performance, these models fail to provide clear explanations for their final results, limiting their interpretability and practical use in real financial decision-making.

\subsection{Financial Time-Series Forecasting}
Modern financial time series forecasting increasingly leverages hybrid architectures. \cite{zhou2021informer} proposed Informer, a transformer variant optimized for time series prediction, which reduces the huge inference cost. For high-frequency trading data, such as the limit order book, \cite{kumar2023deep} designed a reinforcement learning framework with volatility-sensitive rewards. The efficiency of different model topologies has also been explored: some works, such as \cite{hou2021st,you2024multi,pei2025dynamic-icaart}, use a graph-based topology, where stocks are represented as nodes with static and dynamic connections; in contrast, traditional multi-time series models follow a sequential topology, treating each stock as an independent time series without explicitly constructing their relationships. Recent work by \cite{wu2021autoformer} introduced Autoformer, which leverages autocorrelation to decompose market trends and seasonal effects; however, the model remains restricted to single-modal input and will not adapt to sudden market changes (e.g., a black swan event). Finally, most of these models directly use pre-processed data and do not address the complexities of processing multimodal unstructured data.

\subsection{Interpretability}
For industrial applications, in particular, interpretability is a key requirement of financial forecasting models, as commercial vendors and regulators require actionable insights into model decisions. Post hoc explainability tools, such as attention maps \cite{lim2021temporal} and saliency methods \cite{simonyan2013deep}, have been introduced for deep learning models, but the explanations provided lack economic foundations. New deep learning architectures have also been introduced to improve interpretability. For example, to disentangle patterns, \cite{lin2021learning} proposed a Temporal Routing Adaptor with optimal transport, which learns distinct trading patterns and assigns stocks to patterns using dynamic routing. For multimodal settings, \cite{zhou2020domain} designed a Domain-Adaptive Neural Attention Network that aligns news sentiment trends with sector-specific price movements via cross-modal attention; however, although domain adversarial training improves robustness to distribution shifts, interpretability is reduced by masking attribution to specific modalities. 


\begin{figure*}[htbp]
    \centering
    \includegraphics[width=0.75\linewidth]{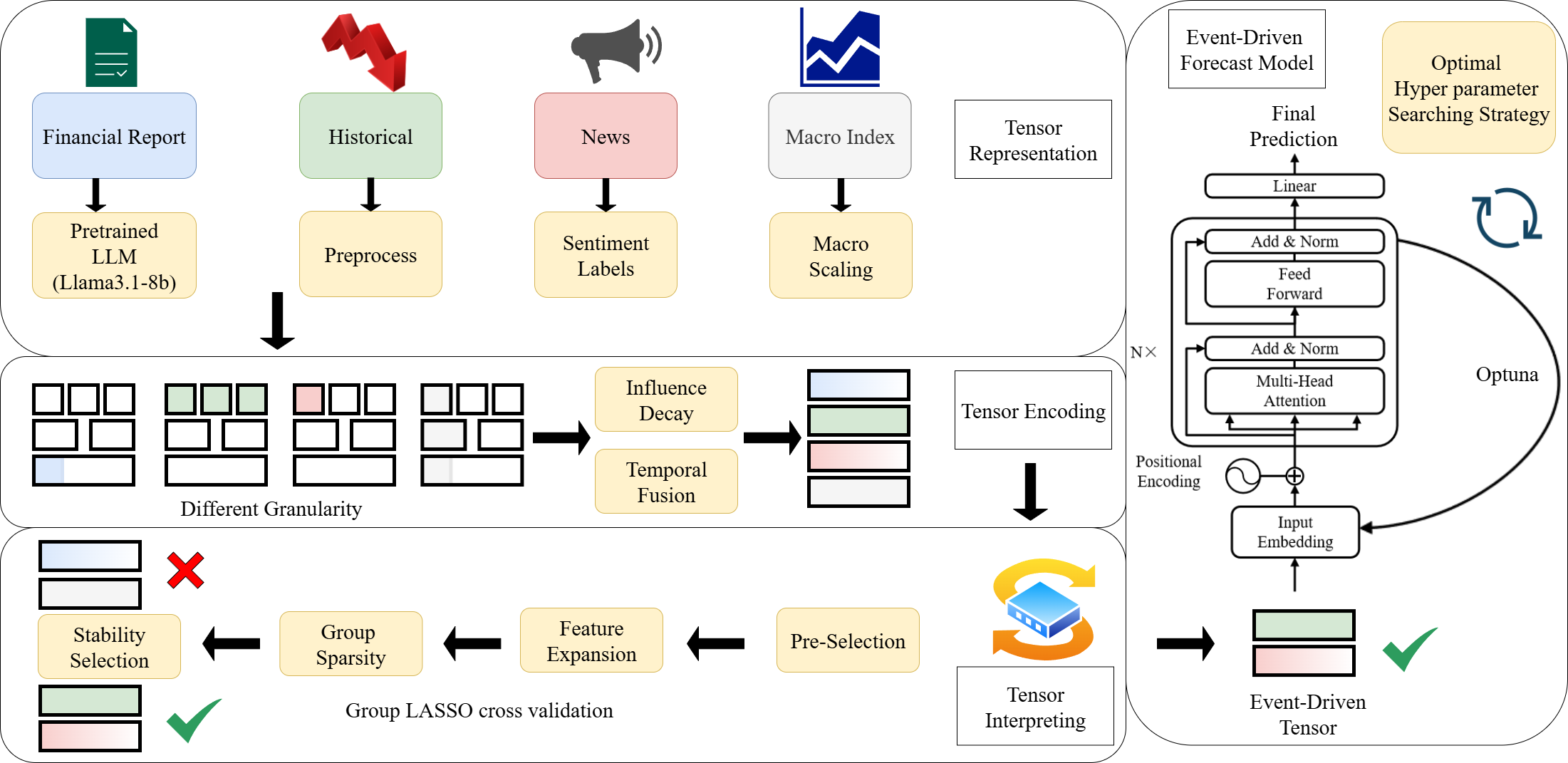}
    \caption{Overview of proposed CMTF. The framework integrates multimodal data (historical data, macro index, news, and financial reports). It employs Tensor Representation (extract tensor representation from unstructured data), Tensor Encoding (scale and preprocess the collected tensor), Tensor Interpretation (select important tensors), and a Transformer-based forecasting model (apply the optimal training scheme).}\label{fig:sentiment-framework}
    \vspace*{5mm} 
\end{figure*}

\section{Preliminary}
\subsection{Notation}
To formalize our methodology, here we define the key notation used for our CMTF model. Let $t$ denote the current time step, and let $T$ be the set of time steps until $t$, so $t \in \{1, ..., T \}$. Let $D$ represent the number of input features after multimodal fusion, and $N$ correspond to the number of target stocks for prediction. Then, the tensor $\mathcal{X} \in \mathbb{R}^{T \times D}$ is the final input tensor containing all encoded features over time, and the model can make predictions for the next day's $(t+1)$ close prices $\hat{P}_{t+1}$ for $N$ stocks, where $i \in \{1, ..., N \}$ is each stock. Finally, $Z^h$, $Z^m$, $Z^n$, and $Z^r$ represent the structured tensors derived from {\em historical data}, {\em macro index}, {\em news}, and {\em financial reports}, respectively. We use this notation consistently throughout the remainder of this paper. 

\subsection{Task Definition}
In this work, we tackle the classification task of financial market forecasting, in which the objective is to predict the direction of the movement of the stock price: whether the stock price will go up or down the next day. At each time step, $t$, we define binary classification labels for the true direction of change. Given that the closing prices in our dataset do not remain the same on two consecutive days, we adopt a straightforward binary classification approach:

\begin{equation}
\text{\em true\_direction}^i_{t+1} = 
\begin{cases}
1, & \text{if } p^i_{t+1} - p^i_{t} > 0 \\
0, & \text{if } p^i_{t+1} - p^i_{t} < 0
\end{cases}
\label{eq:true_direction}
\end{equation}
\noindent
and, similarly, the predicted direction of change:

\begin{equation}
\text{\em pred\_direction}^i_{t+1} = 
\begin{cases}
1, & \text{if } \hat{p}^i_{t+1} - p^i_{t} > 0 \\
0, & \text{if } \hat{p}^i_{t+1} - p^i_{t} < 0
\end{cases}
\label{eq:pred_direction}
\end{equation}
\noindent
where $p^i_{t}$ ($p^i_{t} \in P_{t}$) represents the closing price of stock $i$ at time $t$, and $p^i_{t+1}$ refers to the closing price of stock $i$ at time $t+1$. The classification model aims to predict whether the price of the stock will increase or decrease, with labels $1$ (increase) and 0 (decrease).\footnote{The data contains no instance where daily change in close price is exactly zero.} The objective is to minimize the binary cross-entropy loss between the predicted and true directions.

\section{Methodology}
\noindent
An overview of the CMTF framework is presented in Fig.~\ref{fig:sentiment-framework}. It consists of four components: tensor representation; tensor encoding, tensor interpretation; and a transformer-based forecasting model. Here, we introduce each of these components in detail.

\subsection{Tensor Representation}

\begin{figure}[tb]
    \centering
    \includegraphics[width=0.95\linewidth]{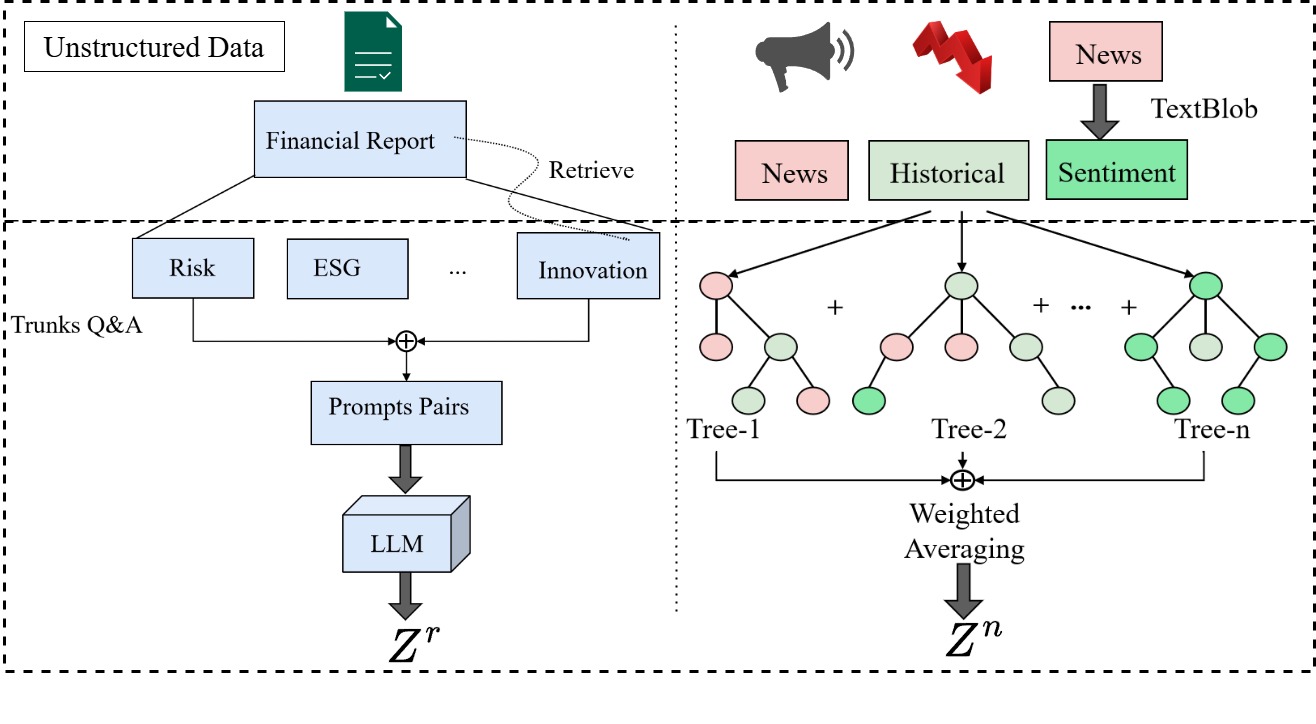}
    \caption{Tensor representation pipeline of financial reports and news.}\label{fig:unstructured}
    \vspace*{10mm} 
\end{figure}

This work uses two types of data: {\em structured data}, consisting of predefined numerical values that can be used directly for model training; and {\em unstructured data}, including textual information, which requires pre-processing before training. Here, we specifically focus on textual data for further processing. At the stage of tensor representation, we aim to transform different data types into tensor representations suitable for further training in Fig.~\ref{fig:unstructured}.

Specifically, we consider two types of unstructured textual data: {\em news} and {\em financial report}. To represent these data, we introduce two complementary approaches: CatBoost for extracting classification tensors $Z^n$ for news; and a Large Language Model (LLM) to generate rating value tensors $Z^r$ for financial reports.

To extract the tensor representation from unstructured news data, we implement a version of CatBoost's gradient-boosting optimized decision trees \cite{prokhorenkova2018catboost}. The loss function shows as:

    \begin{equation}
        \begin{aligned}
     \underset{\theta}{\arg\min} \sum_{m=1}^M \Bigg[ \underbrace{\sum \ell(z^{n}_i, F_{m-1}(\mathbf{x}_i) + f_m(\mathbf{x}_i))}_{\text{Boosted Loss}} \\
     + \underbrace{\lambda ||f_m||^2}_{\text{L2 Reg}} 
     + \underbrace{\gamma(\mathbf{x}_i)}_{\text{Encoding Stabilizer}} \Bigg] 
        \end{aligned}
    \end{equation}

\noindent
where $\mathbf{x}_i$ combines processed text signals and market technicals, including news sentiment, intraday price dynamics, and historical volatility; and $z^{n}_i$ represents the binary classification labels of next day price movements. Both $\mathbf{x}_i$ and $z^{n}_i$ contribute to the output tensor $Z^n$ from unstructured news data. $\theta$ denotes the parameters of the CatBoost model being optimized, while $F_{m-1}(\mathbf{x}_i)$ is the ensemble prediction from the first $m{-}1$ trees, and  $\gamma$ is for penalize.

To transform unstructured Financial report data into a structured rating representation tensor, we employ a Large Language Model (LLM) as an NLP tensor extractor. Specifically, there are two steps.

First, given an input document $U$, the LLM generates a five-dimensional rating vector $R \in \mathbb{R}^{5}$. Second, to ensure compatibility with downstream tasks, we map $R$ into a structured feature space optimized for predictive modeling. This transformation is defined as:  

\begin{equation}
    Z^r = f_{\text{proj}}(f_{\text{rate}}(U))
\end{equation}
  
where $f_{\text{rate}}(\cdot)$ leverages contextual embeddings to infer rating scores, and $Z^r$ represents the final structured representation used for model training and inference. Finally, we obtained four types of tensors for the next stage of encoding: tensors from historical data $Z^h$, macro index $Z^m$, news data $Z^n $, and financial reports $Z^r$.

\subsection{Tensor Encoding}
The influence of a specific event in an event-driven model will usually persist for an extended period, rather than being limited to a single point. Therefore, following the approaches of \cite{ortu2022technical, saragih2024forescating}, we apply a weighted moving average (WMA) to model the decay of influence for data that do not have daily granularity. This assigns higher weights to more recent observations, allowing us to model the diminishing impact of an event over time:

\begin{equation}
{WMA}_t = \frac{\sum_{a=1}^{b} a \cdot S_{t-(b-a)}}{\sum_{a=1}^{b} a} 
\label{eq:wma}
\end{equation}

\noindent Here, $b$ denotes the fixed window size; and $S_{t-(b-a)}$ represents the observation (e.g., news sentiment score) at time $t-a$. This approach allows us to model the extended influence over subsequent $a$ days.

Once the WMA is calculated, we apply this Temporal Fusion (TF) to all tensors:

\begin{equation}
\text{Z}^{h,m,n,r}_{daily} = \text{TF}_{t}({Z}^{h,m,n,r}) \quad
\end{equation}

We then concatenate the resulting tensors to form the final feature set $\mathcal{X} = \text{Concat} (\text{Z}^{h}_{daily},\ldots,\text{Z}^{r}_{daily}) \in \mathbb{R}^{T \times D}$.

\label{subsec:tensor_interp}
To decode cross-modal interactions in financial tensors, we propose an interpretable feature selection framework, combining temporal sparsity and stability analysis. Given input tensor $\mathcal{X} \in \mathbb{R}^{T \times D}$, the pipeline proceeds through four stages:

\paragraph{Correlation-Guided Pre-selection} \label{sssec:corr_preselect}
We first eliminate multicollinear features through mean absolute correlation thresholding:
\begin{equation}
    \Phi_{\text{corr}} = \left\{ d \in [1,D] \, \middle| \, \frac{1}{D-1}\sum_{\substack{d'=1 \\ d' \neq d}}^D |\rho(\mathbf{x}_d, \mathbf{x}_{d'})| < \tau_{\text{corr}} \right\}
\end{equation}
where correlation score $\tau_{\text{corr}}$ is computed from $\mathcal{X}$'s correlation matrix.

\paragraph{Temporal Feature Expansion} \label{sssec:temp_expand}
Next, we construct lagged features to capture delayed market responses through first-order temporal convolution:

\begin{equation}
    \tilde{\mathcal{X}}_{t,d} = \begin{bmatrix} \mathcal{X}_{t^{'},d} \\ \mathcal{X}_{t^{'}-1,d} \end{bmatrix}
    \quad \forall d \in \Phi_{\text{corr}}, \, t^{'} \in \{2,\ldots,T^{'}\}
\end{equation}

\paragraph{Multi-Task Group Sparsity} \label{sssec:group_sparse}
We then solve the convex temporal group LASSO objective \cite{wang2023improved, zhou2010exclusive}:
\begin{equation}
    \min_{W \in \mathbb{R}^{|\Phi_{\text{corr}}| \times N}} \underbrace{\frac{1}{2T^{'}}\|Y - \tilde{\mathcal{X}}W\|_F^2}_{\text{Reconstruction Error}} + \alpha \underbrace{\sum_{d=1}^{|\Phi_{\text{corr}}|} \|W_d\|_2}_{\text{Cross-Target Sparsity}}
    \label{eq:group_lasso}
\end{equation}
where \( Y \) is the matrix of ground truth outputs for target series.\( \| \cdot \|_F \) denotes the Frobenius norm, and \( \frac{1}{2T^{'}} \) normalizes the squared error over the total number of time steps. The group LASSO penalty \( \sum_d \|W_d\|_2 \) encourages sparsity at the feature level across all targets, meaning that only the most relevant features are selected.

\subsection{Tensor Interpretation} 
\begin{figure}[tb]
    \centering
    \includegraphics[width=0.95\linewidth]{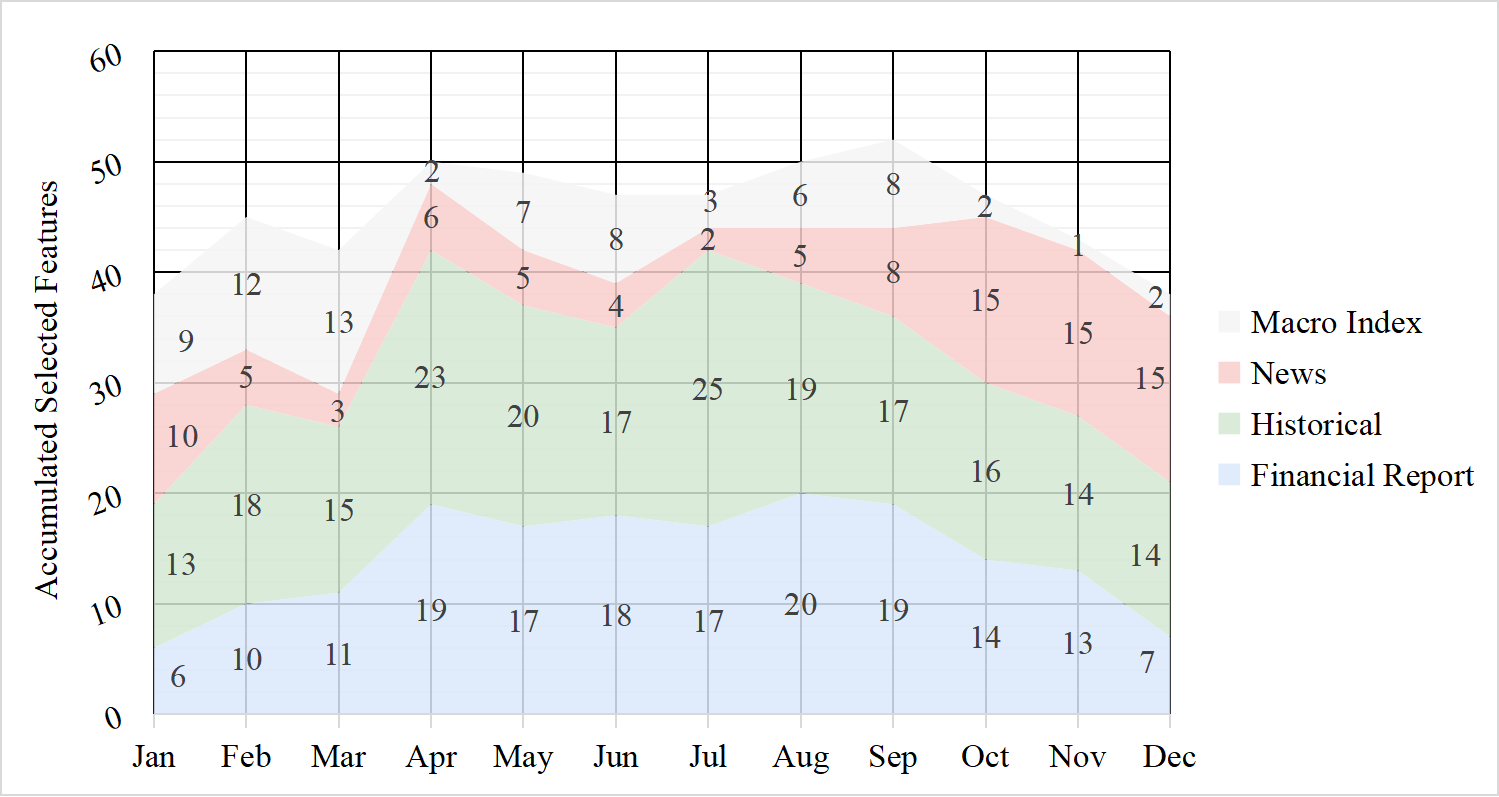}
    \caption{Demonstration of how tensor interpretation accumulates feature values over time. Here, we assume that CMTF undergoes monthly training iterations; a higher label count indicates greater importance for that period.}\label{fig:interpretation}
    \vspace*{10mm} 
\end{figure}

\paragraph{Stability Selection} 
Finally, we retain features with persistent predictive power across temporal folds through majority voting:
\begin{equation}
    \Phi_{\text{final}} = \left\{ d \in \Phi_{\text{corr}} \, \middle| \, \frac{1}{K}\sum_{k=1}^K \mathbb{I}\left(\|W_d^{(k)}\|_2 > 0\right) \geq 0.8 \right\}
    \label{eq:stability}
\end{equation}
where $K$ is the number for temporal splits preserve chronological order in $\mathcal{X}^{'}$, the output tensor which will be used in the final step.

As demonstrated in Fig.~\ref{fig:interpretation}, the relative frequency of each feature value determines its importance. As the training iterates monthly, it identifies key features that are important in specific time windows. This accumulation highlights stable or strongly correlated features, which improves the interpretation of cross-modal interactions over time.

\begin{table*}[tb]
\small
\centering
\caption{Data Summary.}
\label{table:data_introduction}
\resizebox{0.75\textwidth}{!}{%
\renewcommand{\arraystretch}{1.2}
\begin{tabular}{|l|c|c|c|c|c|c|}
\hline
\textbf{Category}    & \textbf{Historical Data} & \multicolumn{3}{c|}{\textbf{Macro index}} & \textbf{News} & \textbf{Financial Reports} \\ \hline
\textbf{Subject}     & From 5 Companies              & \multicolumn{3}{c|}{From 2 Countries}       & From 5 Companies   & From 5 Companies               \\ \hline
\textbf{Subcategory} & Historical Prices        & 1/10Y-Bond-Yield & GDP          & CPI         & Company News & Financial Reports        \\ \hline
\textbf{Detail}      & OHLCV                    & OHLC             & Index        & Value       & Text        & Text                    \\ \hline
\textbf{Granularity} & Daily                    & Daily            & Quarterly& Monthly     & Daily       & Quarterly               \\ \hline
\end{tabular}
}
\vspace*{4mm} 
\end{table*}


\subsection{Event-Driven Forecast Model}

The event-driven forecast model includes a transformer and an optimizer for rapid hyperparameter updates using the filtered feature $\mathcal{X}^{'}$ from the tensor interpretation. 

\paragraph{Transformer}
The encoder in our transformer model consists of several key components, including multi-head attention (MHA), feed-forward layers (FFN), positional encoding (PE), and layer normalization.

Let \( H_l \in \mathbb{R}^{T \times d_{\text{model}}} \) denote the input to the attention layer at encoder layer \( l \), where \( d_{\text{model}} \) is the feature dimension of each token. This input \( H_l \) includes the original feature embedding combined with positional encodings, following standard practice in transformer architectures \cite{vaswani2017attention, devlin2019bert}.

For each attention head \( h \in \{1, \ldots, H\} \), we compute:
\begin{equation}
    Q_h = H_lW_h^Q, \quad K_h = H_lW_h^K, \quad V_h = H_lW_h^V
\end{equation}
\begin{equation}
    \text{Attention}_h = \text{softmax}\left(\frac{Q_hK_h^T}{\sqrt{d_k}}\right)V_h
\end{equation}
\begin{equation}
    \text{MHA}(H_l) = \text{Concat}(\text{Attention}_1,\ldots,\text{Attention}_H)W^O
\end{equation}

\noindent
where \( Q \), \( K \), and \( V \) represent the query, key, and value matrices, respectively. The attention scores are normalized using the softmax function and applied to the value matrix \( V \) to produce the output.

Each encoder layer also contains a position-wise feed-forward network. This consists of two fully connected layers with a ReLU activation function applied between them. The feed-forward network is applied independently to each position in the sequence. The operation can be formally written as:

\begin{equation}
    \text{FFN}(x) = \text{ReLU}(W_1 x + b_1) W_2 + b_2
\end{equation}

\noindent 
where \( W_1 \) and \( W_2 \) are learnable weights, and \( b_1 \) and \( b_2 \) are bias terms. 
Then, the positional encoding function \( PE \) is given by:

\begin{equation}
\text{PE}_{(pos, 2\delta)} = \sin \left( \frac{pos}{10000^{2\delta/d_{\text{model}}}} \right)
\end{equation}
\begin{equation}
\text{PE}_{(pos, 2\delta+1)} = \cos \left( \frac{pos}{10000^{2\delta/d_{\text{model}}}} \right)
\end{equation}

\noindent
where \( pos \) is the position of the token, and \( \delta \) is the dimension index of the positional encoding.

Finally, the output is taken from the last time step of the sequence, and a linear layer is applied to produce the predictions:

\begin{equation}
\hat{P}_{t+1} = \text{Linear}(x_{T})
\end{equation}
\noindent
Here, \( x_{T} \in \mathbb{R}^{d_{\text{model}}} \) refers to the hidden state from the final encoder layer at the last chronological time step, and \( \hat{P}_{t+1} \) represents CMTF's prediction

\begin{figure}[t]
    \centering
    \includegraphics[width=1.0\linewidth]{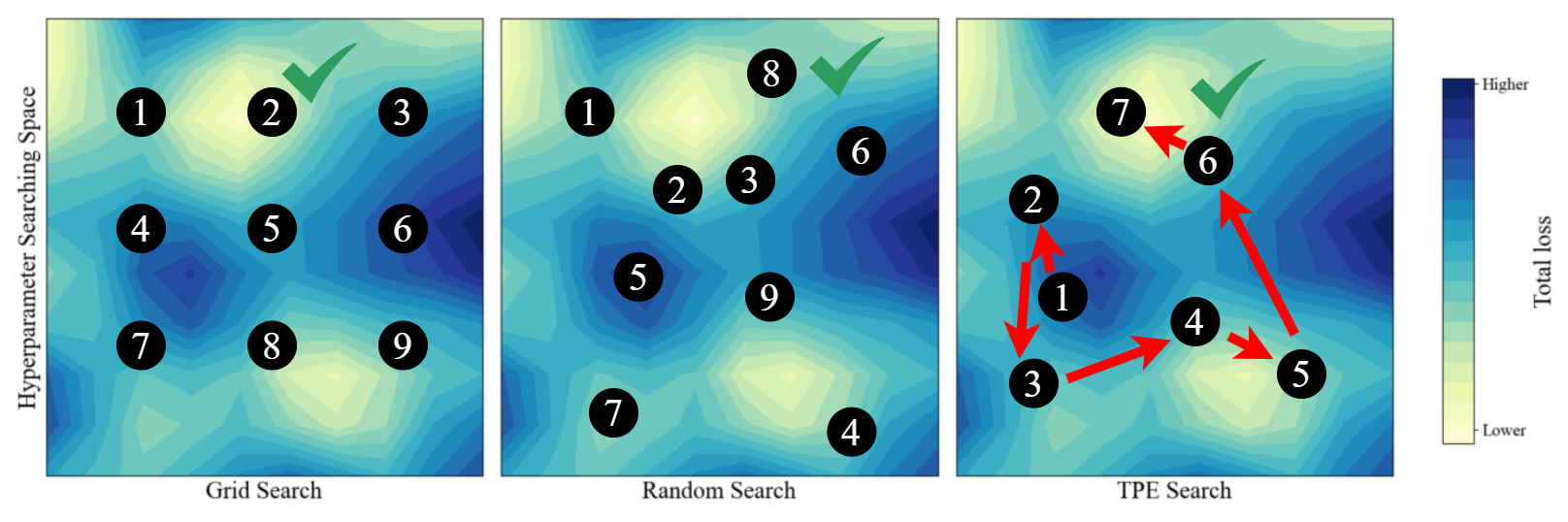}
    \caption{Hyperparameter search strategies over loss landscape: grid search, random search, and tree-structured parzen estimator (TPE).}\label{fig:three_search}
    \vspace*{10mm} 
\end{figure}

\paragraph{Optimizer}
To enable rapid updates and efficient training, we use Optuna \cite{akiba2019optuna} as the optimization framework. Optuna utilizes an asynchronous successive halving algorithm, which is equipped with different estimators to search for the local optimum in the hyperparameter space. The pruning criterion for each trial is defined as:

\begin{equation}
\text{Prune}(t) = \begin{cases}
\text{True}, & \text{if } \frac{r_k}{r_{k-1}} > \gamma^{1/\eta} \\
\text{False}, & \text{otherwise}
\end{cases}
\end{equation}
where \( r_k \) is the trial's intermediate value at step \( k \), \( \eta \) is the reduction factor, and \( \gamma \) is the threshold.

Here, in the CMTF framework, we apply the default estimator, Tree-structured Parzen Estimator (TPE). The TPE estimator models the probability of a set of hyperparameters \( x \) given the value of the objective function \( y \) as:

\begin{equation}
p(x|y) = \frac{\ell(x)}{\ell(x) + g(x)}, \quad y \sim \text{Gamma}(k,\theta)
\end{equation}
\noindent
where \( \ell(x) \) and \( g(x) \) represent the likelihood functions for good and bad hyperparameter configurations, respectively, and \( y \) follows a Gamma distribution with shape \( k \) and scale parameter \( \theta \). As shown in Fig.~\ref{fig:three_search}, TPE demonstrates greater efficiency in identifying locally optimal hyperparameter combinations.



\begin{table*}[htbp]
\footnotesize
\centering
\caption{Data Statistics.}
\label{table:data_statistics}
\resizebox{0.6\textwidth}{!}{%
\renewcommand{\arraystretch}{1.2}
\begin{tabular}{|l|c|c|c|c|c|c|}
\hline
\textbf{Category}         & \textbf{Historical Data} & \multicolumn{3}{c|}{\textbf{Macro index}} & \textbf{News} & \textbf{Financial Reports} \\ \hline
\textbf{Data Type}        & \multicolumn{4}{c|}{\textbf{Structured Data}} & \multicolumn{2}{c|}{\textbf{Unstructured Data}} \\ \hline
\textbf{\# Extracted Structured Features} & \multicolumn{4}{c|}{-}                        & 2 Labels \& 1 Score & 5 Types of Ratings        \\ \hline
\textbf{\# Total Features} & 25 & \multicolumn{3}{c|}{20} & 15 & 25 \\ \hline
\textbf{Time Span}        & \multicolumn{6}{c|}{02/04/2019 – 05/22/2024 (1360 Days)}                             \\ \hline
\textbf{Data Split}       & \multicolumn{6}{c|}{0.6 : 0.2 : 0.2 (Train, Validation, Test)}                          \\ \hline
\textbf{\# Day Split}        & \multicolumn{6}{c|}{804 : 268 : 268 (Train, Validation, Test)}                          \\ \hline
\end{tabular}
}
\vspace*{4mm} 
\end{table*}


\section{Empirical Analysis of CMTF}
In this section, we describe our empirical analysis of CMTF. The analysis is designed to answer three questions:
\begin{description}
    \item[RQ1] How effectively does CMTF forecast financial markets? 
    \item[RQ2] How effective are the individual modules of CMTF?
    \item[RQ3] How does sensitivity to the tensor interpretation module affect performance?
\end{description}

\subsection{Dataset}
Table~\ref{table:data_introduction} summarizes our raw data, and Table~\ref{table:data_statistics} introduce the preprocessed data. The raw data covers 1360 days (02/04/2019--05/22/2024). We integrate structured financial data with unstructured textual sources from five representative UK-headquartered multinational corporations listed in the FTSE 100 index: \textit{Shell}, \textit{Unilever}, \textit{British American Tobacco}, \textit{BP}, and \textit{Diageo}. Macro indexes are chosen from the US and UK to represent the macroeconomics of the world and the target market. 

To train CMTF, the data is partitioned chronologically into distinct splits: 804 training days (02/2019 -- 07/2022), 268 validation days (08/2022 -- 09/2023), and 268 test days (10/2023 -- 05/2024).  The final tensor structure preserves cross-modal interactions between market movements (price), macro-indexes (bond/GDP/CPI), and corporate disclosures (news/reports). For more details, please refer to our GitHub.

\subsection{Configuration}
All experiments were performed on an Nvidia GeForce RTX 4060 laptop GPU with CUDA version 12.6. The lookback window $b$ for the weighted moving average $WMA_t$ is set to 30, while the temporal feature expansion window $T^{'}$ is 90. The five-dimensional rating vector comprises Risk, Market Conditions, Regulation, ESG, and Innovation, with rating scores ranging from 1 to 9, extracted from the pretrained LLM (Llama-3.1-8B). Missing values are handled using linear interpolation for numerical data, while zero-imputation is applied to text embeddings.  

To optimize the Transformer model, we employ Optuna \cite{akiba2019optuna} for efficient hyperparameter tuning, covering both the model architecture and the training parameters.  
For the model architecture, the embedding dimension is selected from $\{32, 64, 128, 256, 512, 1024\}$. The number of attention heads is chosen from $\{2, 4, 8, 16\}$, ensuring divisibility by $d_{\text{model}}$. The model consists of multiple transformer encoder layers, with $\text{num\_layers}$ set from $\{1, 2, 4, 8\}$. The FFN layer dimension is optimized from $\{256, 512, 1024, 2048, 4096\}$.  

For training, the learning rate is searched within $\{$1e-5, 5e-5, 1e-4, 5e-4, 1e-3, 5e-3, 1e-2$\}$. The batch size is chosen from $\{32, 64, 128\}$, balancing computational efficiency and model convergence. The number of training epochs is set from $\{10, 20, 50, 100\}$ to regulate training duration and stability.   
For baselines, ARIMA was configured with automatic order selection, which finds the best combination of ($p$, $d$, $q$) by evaluating multiple models; LSTM used sequential inputs with 50 units, ReLU activation, and 200 training epochs; and SVR used a linear kernel with separate models trained for each target variable, with an average result calculated.

\begin{table}[tb]
\centering
\caption{Classification performance comparison.}
\label{table:classification_comparison}
\resizebox{0.49\textwidth}{!}{%
\begin{tabular}{@{}lccccccc@{}}
\toprule
& Zero & Linear & ARIMA & RF & SVR & LSTM & CMTF \\
\midrule
Precision (\%) $\uparrow$ & 48.71 & 49.33 & 47.13 & 51.54 & 50.10 & 49.49 & 51.04  \\
Recall (\%) $\uparrow$ & 48.86 & 78.21 & 38.58 & 71.10 & 77.16 & 7.41 & 84.88  \\
F1 Score $\uparrow$  & 0.49 & 0.61 & 0.42 & 0.60 & 0.61 & 0.13 & 0.64\\
\bottomrule
\end{tabular}
}
\vspace*{4mm} 
\end{table}

\subsection{Baseline Comparison Models}
We benchmark our framework against various forecasting models that span the methodological spectrum from interpretable linear statistical models to neural network architectures. These are chosen to rigorously test our framework's ability to integrate multimodal data and temporal dynamics beyond conventional approaches. 

\paragraph{Null Model} \leavevmode\\
\textbf{Zero Change:} Price prediction: tomorrow's close will equal today's close: $p_{t+1}=p_t$. Direction prediction: tomorrow's direction will equal today's direction.

\paragraph{Classical Statistical Models} \leavevmode\\
\textbf{Linear Regression:} Models linear relationships between dependent and independent variables by minimizing the sum of squared residuals to fit an optimal hyperplane \cite{weisberg2005applied}. \\
\textbf{ARIMA:} Combines autoregressive (AR), differencing (I), and moving average (MA) components to capture temporal dependencies, trends, and seasonality in non-stationary time series \cite{box2015time}.

\paragraph{Machine Learning Approaches} \leavevmode\\
\textbf{Random Forest:} An ensemble method that aggregates predictions from multiple decorrelated decision trees, reducing overfitting via bootstrap aggregation and feature randomization \cite{breiman2001random}. \\
\textbf{Support Vector Regression (SVR):} Extends support vector machines to regression tasks by mapping inputs to a high-dimensional space and optimizing a margin-sensitive loss function \cite{drucker1996support}.

\paragraph{Deep Learning Architectures} \leavevmode\\
\textbf{LSTM:} A recurrent neural network variant with gating mechanisms (input, output, forget gates) to model long-term sequential dependencies while mitigating vanishing gradients \cite{hochreiter1997long}. \\
\textbf{Encoder-only transformer:} Adapts self-attention \cite{vaswani2017attention} mech\-an\-isms for time series by encoding positional information and temporal relationships, following techniques in \cite{devlin2019bert}.

\subsection{Evaluation Metrics}
Following the approaches taken in previous studies \citep{shobayo2024innovative, singh2021feature, pei2025dynamic-icaart, deng2019knowledge, sawhney2020deep}, we assess our result using Precision, Recall, and F1 score to evaluate model performance. For all three metrics, higher values indicate better model performance. 

\begin{equation}
    \text{Precision} = \frac{\text{TP}}{\text{TP} + \text{FP}}
\end{equation}

\begin{equation}
    \text{Recall} = \frac{\text{TP}}{\text{TP} + \text{FN}}
\end{equation}

\begin{equation}
    \text{F1 Score} = 2 \times \frac{\text{Precision} \times \text{Recall}}{\text{Precision} + \text{Recall}}
\end{equation}

In our classification scheme: true positive (TP) indicates we correctly predicted an increase in price; true negative (TN) indicates we correctly predicted a decrease in price; false positive (FP) indicates we incorrectly predicted a price increase when the price decreased; and false negative (FN) indicates we incorrectly predicted a price decrease when the price increased. 
Note that we focus only on classification metrics and deliberately avoid using error metrics that are often applied for regression tasks. This is because a zero change model ($p_{t+1}=p_t$) will return a low root mean squared error \ifnum\APPENDIX=1{(RMSE)}\fi or mean absolute percentage error\ifnum\APPENDIX=1{ (MAPE)}\fi. \ifnum\APPENDIX=1{Appendix~\ref{app:regression} shows that the zero change model has the lowest RMSE and MAPE of all models analysed.}\fi 

\begin{table*}[tbh]
\centering
\caption{Ablation study on CMTF with different configurations (\textit{+}/\textit{-}, \textit{I}/\textit{N}/\textit{R}), 
denoting with/without tensor interpreting (I), news (N), and financial reports (R).}
\label{table:ablation_classification}
\resizebox{0.5\textwidth}{!}{%
\renewcommand{\arraystretch}{1.05}
\begin{tabular}{@{}lcccccccc@{}}
\toprule
Metric & \multicolumn{4}{c}{\textit{+ I}} & \multicolumn{4}{c}{\textit{- I}} \\
\cmidrule(lr){2-5} \cmidrule(lr){6-9}
& \textit{+N+R} & \textit{+N-R} & \textit{-N+R} & \textit{-N-R} 
& \textit{+N+R} & \textit{+N-R} & \textit{-N+R} & \textit{-N-R} \\
\midrule
Precision (\%) $\uparrow$ & 51.44 & 49.40 & 49.69 & 49.57 & 49.91 & 50.18 & 51.29 & 49.79 \\
Recall (\%) $\uparrow$ & 45.51 & 49.40 & 72.01 & 69.46 & 80.09 & 60.93 & 65.42 & 72.16 \\
F1 Score $\uparrow$ & 0.48 & 0.49 & 0.59 & 0.58 & 0.61 & 0.55 & 0.58 & 0.59 \\
\bottomrule
\end{tabular}
}
\vspace*{5mm} 
\end{table*}

\begin{table*}[tbh]
\centering
\caption{Ablation study comparing base methods with Tensor Interpreting (\textit{+ I}).}
\label{table:I_ablation_classification}
\resizebox{0.7\textwidth}{!}{%
\begin{tabular}{@{}lcccccccccccc@{}}
\toprule
\multirow{2}{*}{Metric} & \multicolumn{2}{c}{Linear Regression} & \multicolumn{2}{c}{ARIMA} & \multicolumn{2}{c}{Random Forest} & \multicolumn{2}{c}{SVR} & \multicolumn{2}{c}{LSTM} & \multicolumn{2}{c}{Transformer} \\
\cmidrule(lr){2-3} \cmidrule(lr){4-5} \cmidrule(lr){6-7} \cmidrule(lr){8-9} \cmidrule(lr){10-11} \cmidrule(l){12-13}
& Base & +I & Base & +I & Base & +I & Base & +I & Base & +I & Base & +I \\
\midrule
Precision (\%) $\uparrow$ & 49.55 & 49.22 & 47.13 & 47.13 & 50.41 & 49.84 & 49.61 & 51.62 & 50.40 & 48.24 & 51.29 & 49.69 \\
Recall (\%) $\uparrow$    & 57.79 & 61.72 & 38.58 & 38.58 & 75.19 & 69.59 & 76.25 & 67.62 & 19.21 & 45.54 & 65.42 & 72.01 \\
F1 Score $\uparrow$       & 0.53  & 0.55  & 0.42  & 0.42  & 0.60  & 0.58  & 0.60  & 0.59  & 0.28  & 0.47  & 0.58  & 0.59 \\
\bottomrule
\end{tabular}%
}
\vspace*{5mm} 
\end{table*}

\subsection{RQ1 Performance Comparison}
\noindent
Table \ref{table:classification_comparison} presents a comprehensive performance evaluation of our proposed CMTF against different baselines. To test the effectiveness of our framework, we do not enable the Tensor Representation module: we only include Tensor Encoding, Tensor Interpreting, and Transformer forecasting in the classification settings.
In classification, our proposed CMTF framework exhibits the highest recall of 84.88\% and an F1-score of 0.64, outperforming all baselines. This highlights its strength in capturing sequential dependencies and leveraging multimodal data sources to improve predictive accuracy. Random Forest achieves an F1-score of 0.60, indicating its effectiveness in feature selection and ensemble learning, though its recall 71.10\% remains lower than CMTF.


The experimental results reveal a key advantage of CMTF: its ability to effectively integrate heterogeneous data sources while maintaining strong predictive capabilities. Unlike traditional models that rely on a single data modality, CMTF exploits tensor factorization to capture cross-modal dependencies, leading to superior classification performance.

\subsection{RQ2 Ablation Study}

To further investigate the contribution of different components in our CMTF framework, we performed an ablation study with various configurations, as shown in Table~\ref{table:ablation_classification}. Here, Macro Scaling is enabled by default. The experiments analyze the impact of three key factors: Tensor Interpretation module (\textit{I}), news data (\textit{N}), and financial reports (\textit{R}).
For classification, the highest recall 80.09\% and best F1-score 0.61 occur when Tensor Interpretation is disabled (\textit{-I}) but both News and Financial Reports are included (\textit{+N, +R}). This suggests that textual modalities are crucial for predicting market movement direction, likely due to their ability to capture sentiment and fundamental shifts.

\subsection{RQ3 Module Sensitivity}

To understand the impact of Tensor Interpreting (\textit{+I}), we conduct an ablation study to compare baseline methods with and without this component.
From Table \ref{table:I_ablation_classification}, we find that the impact of \textit{+I} is nuanced. Precision remains relatively stable across models, indicating that Tensor Interpreting does not compromise class separability. In contract, recall exhibits noticeable improvements, particularly for Transformer and LSTM, suggesting that \textit{+I} aids in identifying more relevant patterns for classification.
The results confirm that Tensor Interpretation enhances  performance, particularly for models that rely heavily on feature transformations (e.g., SVR, LSTM, Transformer). 

\section{Discussion}
Although we have evaluated our framework across multiple time series forecasting models, several challenges remain. One key limitation is the absence of a publicly available standardized dataset for multimodal and cross-modal financial forecasting. For example, in \cite{mandal2024enhancing}, they used data from \cite{xu2018stock}, which includes only tweets and historical prices, limiting its applicability to broader multimodal scenarios. Although \cite{lim2021temporal} discussed methods for handling different data granularities, it does not address the integration of unstructured data, such as textual information, into forecasting models. Another major challenge is data privacy. In a previous study \cite{cheng2022financial}, they compiled a diverse dataset that included financial events, news, historical prices, and knowledge graph data. However, due to privacy concerns, the dataset was not made publicly available, restricting reproducibility and benchmarking opportunities for future research.

Given these constraints, our evaluation focuses primarily on comparative baseline experiments after the feature engineering stage. Addressing these challenges, either by developing open multimodal financial datasets or by refining privacy-preserving data-sharing mechanisms, would be a crucial step forward in the future.

Future work for CMTF could explore the correlation between financial reports (R) and news (N), as their sentiment may be inherently linked. While current results show combined effects, analyzing their individual contributions may uncover deeper interactions and improve the interpretability of the CMTF framework. Additionally, the strong performance of simpler models such as SVR suggests that unstructured data may be less relevant for next-day price level prediction. This points to the need for task-specific model selection, where increased complexity is justified only when it adds meaningful predictive value. Exploring when and why simpler models outperform could provide valuable insights into the limits and optimal use cases of multi-modal approaches like CMTF.

\section{Conclusion}
We introduce Cross-Modal Temporal Fusion (CMTF), a transformer-based deep learning framework for financial market forecasting. To effectively capture the interactions between historical price trends, macro indexes, and textual financial data, CMTF incorporates specialized components. These include: (1) an attention-based cross-modal fusion mechanism that dynamically weighs the contribution of different modalities, (2) a tensor interpretation module to extract relevant cross-modal features, and (3) an auto-training scheme to streamline model iteration and optimization. Using real-world financial data sets, we demonstrate that CMTF outperforms all baselines on the classification task. Lastly, we explore the interpretability of our model, highlighting how CMTF can (i) analyze the relative importance of different modality data and (ii) adapt to evolving market dynamics through its feature selection mechanisms. For industrial users, CMTF is more than just a financial market forecasting model; it serves as a robust framework to efficiently handle multimodal data.



\begin{ack}
This work was supported by UK Research and Innovation (UKRI) Innovate UK grant number 10094067: Stratlib.AI - A trusted machine learning platform for asset and credit managers.
\end{ack}



\ifnum\APPENDIX=1{
\appendix
\section{Appendix: Regression Analysis}\label{app:regression}
Here we provide regression analysis of the models. The regression task focuses on predicting the next day's closing price for all $N$ stocks. Given the input tensor $\mathcal{X} \in \mathbb{R}^{T \times D}$, the model aims to predict the future stock price at time $t+1$. The objective is to minimize the mean squared error between the predicted and actual stock prices:

\begin{equation}
\hat{P}_{t+1} = f(\mathcal{X}_T)
\end{equation}
\noindent
where $\hat{P}_{t+1}$ is a vector of predicted multi-stock closing prices at time $(t+1)$, and $f$ represents the learned function. The goal is to accurately forecast the next closing price of all $N$ stocks.

We use the widely adopted error-based metrics RMSE and MAPE, where lower values indicate better predictive accuracy. RMSE and MAPE are defined as:

\begin{equation}
    \text{RMSE} = \sqrt{\frac{1}{n} \sum_{i=1}^{n} (\hat{p}^{i}_{t+1} - {p}^{i}_{t+1})^2}
\end{equation}

\begin{equation}
    \text{MAPE} = \frac{1}{n} \sum_{i=1}^{n} \left| \frac{\hat{p}^{i}_{t+1} - {p}^{i}_{t+1}}{{p}^{i}_{t+1}} \right| \times 100
\end{equation}  

Results presented in Table~\ref{table:regression_comparison} show that the Zero Change model has the lowest RMSE and MAPE. Despite this, the Zero Change model has very poor classification accuracy, with precision and recall both less than 49\% (see Table~\ref{table:classification_comparison}). This demonstrates that RMSE and MAPE are unsuitable evaluation metrics for models predicting tomorrow's price movement.

\begin{table}[tbh]
\small
\centering
\caption{Performance comparison for regression task across baselines and CMTF.}
\label{table:regression_comparison}
\resizebox{0.45\textwidth}{!}{%
\begin{tabular}{@{}lccccccc@{}}
\toprule
 & Zero & Linear & ARIMA & RF & SVR & LSTM & CMTF \\
\midrule
RMSE $\downarrow$ & 1.07 & 1.31 & 5.54 & 2.79 & 1.21 & 9.25 & 3.35  \\
MAPE (\%) $\downarrow$ & 2.05 & 3.72 & 20.91 & 9.08 & 3.53 & 33.78 & 10.04  \\
\bottomrule
\end{tabular}
}
\end{table}
}\fi

\bibliography{ref}

\end{document}